# Mining Clinical Notes for Physical Rehabilitation Exercise Information: Natural Language Processing Algorithm Development and Validation Study

Sonish Sivarajkumar, MS[1]

Fengyi Gao, MS[2]

Parker E. Denny, DPT[4]

Bayan M. Aldhahwani, PT, MS[4,6]

Shyam Visweswaran, MD, PhD[1,3,5]

Allyn Bove, DPT, PhD[4]

Yanshan Wang, PhD[1,2,3,5]

[1]Intelligent Systems Program, University of Pittsburgh, Pittsburgh, PA

[2]Department of Health Information Management, University of Pittsburgh, Pittsburgh, PA

[3]Department of Biomedical Informatics, University of Pittsburgh, Pittsburgh, PA

[4]Department of Physical Therapy, University of Pittsburgh, Pittsburgh, PA

[5]Clinical and Translational Science Institute, University of Pittsburgh, Pittsburgh, PA

[6]Department of Physical Therapy, Umm Al-Qura University, Makkah, Saudi Arabia

Correspondence:

Yanshan Wang

yanshan.wang@pitt.edu

# Abstract

**Background:** Post-stroke patient rehabilitation requires precise, personalized treatment plans. Natural Language Processing (NLP) offers potential to extract valuable exercise information from clinical notes, aiding in the development of more effective rehabilitation strategies.

**Objective:** This study aims to develop and evaluate a variety of NLP algorithms to extract and categorize physical rehabilitation exercise information from the clinical notes of post-stroke patients treated at the University of Pittsburgh Medical Center.

**Methods:** A cohort of 13,605 patients diagnosed with stroke was identified, and their clinical notes containing rehabilitation therapy notes were retrieved. A comprehensive clinical ontology was created to represent various aspects of physical rehabilitation exercises. State-of-the-art NLP algorithms were then developed and compared, including rule-based, machine learning-based algorithms (Support Vector Machine, Linear Regression, Gradient Boosting, and AdaBoost), and large language model (LLM)-based algorithms (ChatGPT). The study focused on key performance metrics, particularly F1 scores, to evaluate algorithm effectiveness.

**Results:** Analysis was conducted on a dataset comprising 23,724 notes with detailed demographic and clinical characteristics. The rule-based NLP algorithm demonstrated superior performance in most areas, particularly in detecting the 'Right Side' location with an F1 score of 0.975, outperforming Gradient Boosting by 0.063. Gradient Boosting excelled in 'Lower Extremity' location detection (F1 score: 0.978), surpassing rule-based NLP by 0.023. It also showed notable performance in 'Passive Range of Motion' with an F1 score of 0.970, a 0.032 improvement over rule-based NLP. The rule-based algorithm efficiently handled 'Duration', 'Sets', and 'Reps' with F1 scores up to 0.65. LLM-based NLP, particularly ChatGPT with few-shot prompts, achieved high recall but generally lower precision and F1 scores. However, it notably excelled in 'Backward

Plane' motion detection, achieving an F1 score of 0.846, surpassing the rule-based algorithm's 0.720.

**Conclusions:** The study successfully developed and evaluated multiple NLP algorithms, revealing the strengths and weaknesses of each in extracting physical rehabilitation exercise information from clinical notes. The detailed ontology and the robust performance of the rule-based and Gradient Boosting algorithms demonstrate significant potential for enhancing precision rehabilitation. These findings contribute to the ongoing efforts to integrate advanced NLP techniques into healthcare, moving towards predictive models that can recommend personalized rehabilitation treatments for optimal patient outcomes.

## Introduction

Precision medicine is a promising field of research that aims to provide personalized treatment plans for patients[1]. Recent years have seen a rise in interest in this field, as advances in machine learning and data collection techniques have greatly facilitated this research[2]. However, the principles of precision medicine have primarily been applied to development of medications, and relatively little research has been conducted on their applications in other areas[3]. For instance, although rehabilitation clinics require individualized treatment procedures for patients, little research has been conducted on methods that use data analysis and machine learning to facilitate the design of such procedures[4]. Although the application of precision medicine to physical therapy has proven effective in improving the health of patients, current methods of creating personalized treatments rarely use automated approaches to facilitate decision support[5]. Thus, there is a need for tools to assist in the development of personalized treatments in physical therapy[6]. In the treatment of post-stroke patients, the lack of decision support tools is especially pronounced, as the available treatments for this condition have not led to consistent outcomes across patient populations[7].

To develop decision support tools for the design of precision rehabilitation treatments for post-stroke patients, it would be necessary to utilize electronic health record (EHR) data to develop a predictive model of existing treatment options and their impact on patient outcomes[8]. However, physical therapy procedures are typically described in unstructured clinical notes, meaning that simple data extraction methods such as database queries cannot be applied to obtain sufficient information. Additionally, the language used to describe these procedures can differ between clinicians, locations, and periods[9]. More advanced natural language processing (NLP) algorithms are required to extract this information from clinical notes, but such a method has not yet been developed for this application.

In this paper, we aim to develop and evaluate NLP algorithms to extracting physical rehabilitation exercise information from the clinical notes in the EHR. Our specific contributions are:

1) We created a novel and comprehensive clinical ontology to represent physical rehabilitation exercise information, which includes type of motion, side of body, location on body, plane of motion, duration, information on sets and reps, exercise purpose, exercise type, and body position.
2) We developed and compared a variety of NLP algorithms leveraging the state-of-the-art techniques, including rule-based NLP algorithms, machine learning-based NLP algorithms (i.e., Support Vector Machine, Linear Regression, Gradient Boosting, and AdaBoost), and large language model-based NLP algorithms (i.e., ChatGPT) for the extraction of physical rehabilitation exercise from clinical notes. We are among the first to evaluate the capabilities of ChatGPT in extracting useful information from clinical notes.

# Methods

*Figure 1: Flowchart illustrating the data flow throughout the study*

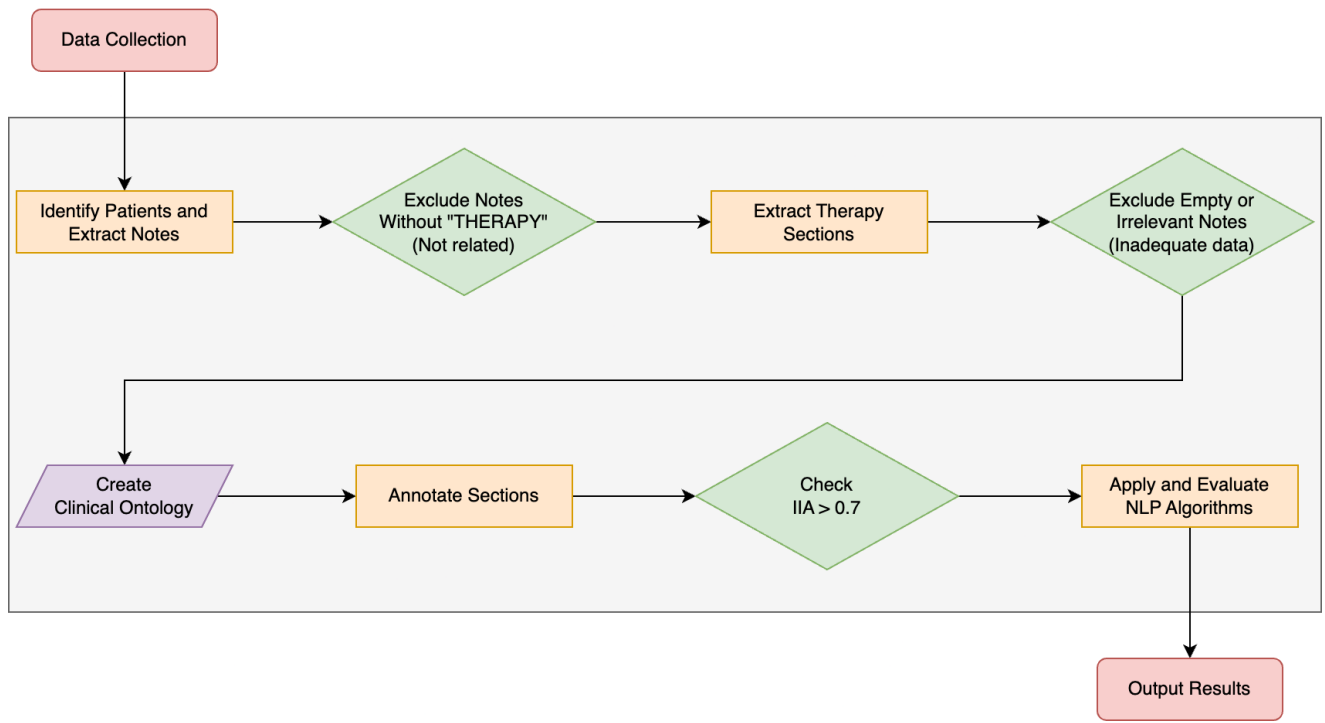

Figure 1 illustrates the data flow and the various stages of the research process. Each of these stages will be described in detail in the following sections.

## Data Collection

The study identified a cohort of patients diagnosed with stroke between January 1st, 2016 and December 31st, 2016 at UPMC. For these patients, clinical encounter notes created between January 1st, 2016 and December 31st, 2018 were extracted from the institutional data warehouse.

The study was approved by the University of Pittsburgh's Institutional Review Board (IRB #21040204). Table 1 provides demographic characteristics of the patients included in this dataset:

Table 1: Demographic information of patients included in the unfiltered dataset

| Demographics | Total: 13,605 |
|---|---|
| Mean Age (std) | 75 years (16) |
| Gender | |
| Female | 6,931 (51%) |
| Male | 6,673 (49%) |
| Race | |
| White | 11,661 (86%) |
| Black | 1,325 (9.7%) |
| Asian | 64 (0.5%) |
| Other | 153 (1.1%) |
| Not Specified | 402 (3.0%) |
| Ethnicity | |
| Hispanic or Latinx | 64 (0.5%) |
| Not Hispanic or Latinx | 12,471 (92%) |
| Not Specified | 984 (7.2%) |

**Clinical Ontology for Physical Rehabilitation Exercise**

To determine the relevance and hierarchy of extracted information, we developed a clinical ontology consisting of nine categories of concepts relating to exercise descriptions, informed by consultation with clinical experts in the field of physical rehabilitation. In developing our clinical ontology, we also consulted established frameworks like the International Classification of

Functioning, Disability, and Health (ICF)[10] and the Systematized Nomenclature of Medicine -- Clinical Terms (SNOMED CT)[11]. These comprehensive systems offered valuable insights into the structuring and categorization of health-related concepts, which we adapted for the specific context of physical rehabilitation exercises. Additionally, our ontology incorporates principles from the Unified Medical Language System (UMLS)[12] to ensure compatibility and interoperability with other healthcare informatics systems.

Each category was given a set of values, as well as examples of how those values might be expressed in clinical notes. The categories are as follows: type of motion, side of body, location on body, plane of motion, duration, information on sets and reps, exercise purpose, exercise type, and body position. The ontology also includes examples of indications that the mentioned exercise was not performed during the visit corresponding to the clinical note. This ontology was used to inform both the structure of the annotations and the methods used to extract relevant documents from the dataset.

The ontology reflects the complexity and nuance of physical rehabilitation exercises by incorporating terms and categories that are sensitive to the variations and specificities observed in clinical settings. This approach ensures that the ontology not only represents the theoretical model of rehabilitation exercises but also aligns with the practical, real-world application and documentation by healthcare professionals. Table 2 displays the nine categories for three exercise descriptions(Performed In-Office, Home Exercise Program, and Not Performed), with sets and reps split into separate rows and including negations and out-of-office exercises at the bottom.

Table 2: Summary of the clinical ontology used for annotations

| Category | Data Type | Concepts |
| --- | --- | --- |

| Exercise Description | Enumerated | Performed In-Office, Home Exercise Program, Not Performed |
|---|---|---|
| Type of Motion | Enumerated | Range of Motion (ROM), Active ROM, Active-Assisted ROM, Passive ROM |
| Side of Body | Enumerated | Right, Left, Bilateral, Unilateral, Contralateral, Ipsilateral |
| Location on Body | Enumerated | Upper Extremity (Arms), Lower Extremity (Legs), Hip, Thigh, Knee, Ankle, Foot, Heel, Toe, Shoulder, Scapula, Elbow, Forearm, Wrist, Hand, Thumb, Head, Neck, Chest, Abdomen, Lower Back |
| Plane of Motion | Enumerated | Flexion, Extension, Abduction, Adduction, Internal Rotation, External Rotation, Lateral Flexion, Horizontal Abduction, Horizontal Adduction, Protraction, Retraction, Elevation, Depression, Inversion, Eversion, Pronation, Supination, Plantarflexion, Dorsiflexion, Radial Deviation, Ulnar Deviation, Upward Rotation, Downward Rotation, Opposition, Forward, Backward, Lateral, Medial, Scaption, Rotation, Closure, Clockwise, Counterclockwise, Distraction, All Planes, Anterior, Posterior, Horizontal, Vertical, Diagonal, Gravity Elimination |
| Duration (seconds) | Integer | N/A |
| Number of Sets | Integer | N/A |
| Number of Reps | Integer | N/A |
| Exercise Purpose | Enumerated | Strength, Fine motor, Motor control, Perception, Simulated, Power, Endurance, Joint Mobility, Joint Alignment, Muscle Flexibility, Cardio, Pulmonary, Agility, Vestibular |
| Exercise Type | Enumerated | Upper Extremity Strength, Lower Extremity Strength, Trunk/Core Strength, Scapular Strength, Range of Motion, Flexibility/Mobility, Balance/Vestibular, Gait Training, |

|  |  | Cardio/Aerobic, Functional Mobility |
|---|---|---|
| Body Position | Binary | Weight Bearing<br>Non-Weight bearing |
| Negation / Hypothetical | Binary | Held / Not Performed<br>Home Exercise Program |

**Preprocessing and Section Extraction**

Physical therapeutic procedures were usually documented in the section "THERAPY". Therefore, we first filtered out the notes that did not contain a physical therapy visit by excluding files whose names lacked the string "THERAPY." From the resulting set of files, the section on therapeutic procedures was extracted using a regular expression, if such a section existed. This resulted in a total of 23,724 notes, some of which were empty or lacked pertinent information.

The method of section extraction has a few minor limitations. Because the regular expression used to locate these sections assumes a structure in the notes that is not always present, it is possible that a file may contain additional text from other sections of the original note in rare instances. All sections utilized in the creation of the gold standard labels were manually examined to ensure the absence of these errors. It is also possible that some therapeutic procedures sections are completely omitted from the note due to copy-and-paste errors made by their authors.

Because many of the extracted sections were very brief or lacked relevant information, we developed a method to create a more robust set of sections by extracting keywords. Initially, concepts were organized into nine categories based on the clinical ontology. Each category was then assigned a list of keywords. A section was considered to mention a category if it contained at least one of the keywords. Consequently, each section was assigned a score between 0 and 9 based on the number of categories mentioned. All sections with a score of 9 and a random selection of notes with a score of 8 were extracted to generate 300 enriched sections that were anticipated to be relatively dense in information. In addition, 300 random sections were selected, excluding those with a length of fewer than 200 characters in order to reduce the likelihood of omissions.

**Gold Standard Dataset Creation**

Three students majoring in rehabilitation or health informatics were recruited as annotators to create gold-standard labels. Each annotator was given a set of guidelines on how to label sections and was told to refer to the clinical ontology for examples of each concept to label. Instructions were given to label explicit mentions of each concept, and inferences were only to be made when specified. For example, the concepts under the categories *exercise type* and *positioning* were each given several common keywords that indicate exercises that relate to them. The annotators were given identical batches of 20 randomly selected sections to annotate, and the Inter-Annotator Agreement (IAA) was calculated using Fleis's Kappa. This process was repeated for a total of three batches, after which all three annotators achieved an IAA greater than 0.7. Throughout this process, the annotation guidelines were revised, and the structure of the labels was finalized. Once sufficient agreement was reached, 50 sections from the enriched set and 50 more from the random set were given to each annotator, totaling 300 distinct annotated sections. These sections were then split randomly into a training set consisting of 125 sections from each of the original sets and a test set consisting of the remaining 50 sections. The details of this corpus is included in Table 3, which outlines the total word count, the number of distinct words, and two examples of the data.

Table 3: Summary of the annotated corpus

| Metric | Value |
|--------|-------|
|        |       |
| **Total Words** | 74104 |
| **Total Distinct Words** | 2371 |
| **Example 1** | 3: LAQ B-3x10 3 #  .7 M/S INC GAIT SPEED TO .7 M/S INC GAIT SPEED TO .7 M/S INC GAIT SPEED TO .7 M/S   Functional Goal 3 (Long Term)  INC GAIT SPEED TO .8 M/S INC GAIT SPEED TO .8 M/S INC GAIT SPEED TO .8 M/S INC GAIT SPEED TO .8 M/S INC GAIT SPEED TO .8 M/S   Pain (Short Term) DEC PAIN AT ITS WORST TO '5' DEC PAIN AT ITS WORST TO '5' DEC PAIN AT ITS WORST TO '5' DEC PAIN AT ITS WORST TO '5' DEC PAIN AT ITS WORST TO '5' DEC PAIN AT ITS WORST TO '5'   Pain (Long Term) DEC PAIN AT ITS WORST TO  '3' DEC PAIN AT ITS WORST TO  '3' DEC PAIN AT ITS WORST TO  '3' DEC PAIN AT ITS WORST TO  '3' DEC PAIN AT ITS WORST TO  '3' DEC  PAIN  AT  ITS  WORST  TO  '3'   Strength  (Short Term) INC STRENGTH [PERSONALNAME] FLEXION TO 4/5 , ABD TO 4/5, ER TO 4/5 INC STRENGTH [PERSONALNAME] FLEXION TO 4/5 , ABD TO 4/5, ER TO 4/5 INC STRENGTH [PERSONALNAME] FLEXION TO 4/5 , ABD TO 4/5, ER TO 4/5 INC STRENGTH [PERSONALNAME] FLEXION TO 4/5 , ABD TO 4/5, ER TO 4/5 INC STRENGTH [PERSONALNAME] FLEXION TO 4/5 , ABD TO 4/5, ER TO 4/5 INC STRENGTH [PERSONALNAME] FLEXION TO 4/5 , ABD TO 4/5,  ER  TO  4/5   Strength  (Long  Term)  INC  STRENGTH [PERSONALNAME] ER / IR TO 4+/5 INC STRENGTH [PERSONALNAME] ER / IR TO 4+/5 INC STRENGTH [PERSONALNAME] ER / IR TO 4+/5 INC STRENGTH [PERSONALNAME] ER / IR TO 4+/5 INC STRENGTH [PERSONALNAME] ER / IR TO 4+/5 INC STRENGTH [PERSONALNAME] ER / IR  TO 4+/5   ROM (Short Term) INC ER/90  ABD TO 60, IR/90  TO 55 INC ER/90  ABD TO 60, IR/90  TO 55 INC ER/90  ABD TO 60, IR/90  TO 55 INC ER/90  ABD TO 60, IR/90  TO 55 INC ER/90  ABD TO 60, |
|  | IR/90 TO 55 INC ER/90 ABD TO 60, IR/90 TO 55  Other (Short Term) INC ACTIVE L UE ELEVATION TO 90  INC ACTIVE L UE ELEVATION TO 90  INC ACTIVE L UE ELEVATION TO 90  INC ACTIVE L UE ELEVATION TO 90  INC ACTIVE L UE ELEVATION TO 90  INC ACTIVE L UE ELEVATION TO 90   Other (Long Term) INC ACTIVE L UE ELEVATION TO 110  INC ACTIVE L UE ELEVATION TO 110  INC ACTIVE L UE ELEVATION TO 110  INC ACTIVE L UE |
| **Example 2** | [ADDRESS]: . 5-10 deg; PROM of shoulder flex and abd + 10-15 deg increased AROM of R shoulder flex and abd, elbow flex + 5-10 deg; PROM of shoulder flex and abd + 10-15 deg increased AROM of R wrist ext 5-10 deg   ROM (Long Term) increased AROM of R shoulder flex and abd, elbow flex + 10-15 deg; PROM of shoulder flex and abd + 15-20 deg increased AROM of R shoulder flex and abd, elbow flex + 10-15 deg; PROM of shoulder flex and abd + 15-20 deg increased AROM of R wrist ext 10-15 deg; shoulder flex, abd, ER 10-15 deg   Other Goal (Short Term) facilitate R stroke recovery stage on CMA arm to 5, and hand to 5 facilitate R stroke recovery stage on CMA arm to 5, and hand to 5   Other Goal (Long Term) facilitate R stroke recovery stage on CMA arm to 6, and hand to 6 facilitate R stroke recovery stage on CMA arm to 6, and hand to 6 facilitate R stroke recovery stage on CMA arm to 6, and hand to 6   Other Goal (Short Term) increase use RUE for ADL up to 50% on CAHAI = 25/49 increase use RUE for ADL up to 60% on CAHAI = 30/49 increase use RUE for ADL up to 70% on CAHAI = 35/49   Other Goal (Long Term) increase use RUE for ADL up to 60% on CAHAI = 30/49 increase use RUE for ADL up to 70% on CAHAI = 35/49 increase use RUE for ADL up to 80% on CAHAI = 40/49   Become Independent with Home Program? Yes Yes Yes" |

**Rule-based NLP**

The first NLP method we developed was a named entity recognition (NER) algorithm using MedTagger, which is a software that uses rule-based methods to segment documents and extract named entity information with regular expressions[13]. We used this tool to detect the categories outlined in the ontology by creating explainable rules to extract the physical rehabilitation exercise information and compare against the gold standard labels. For each rule defined in the algorithm, MedTagger identified spans of text that matched the expression as well as the corresponding category and concept predicted for that text. We initiated the rules using simple keywords in the clinical ontology as defined in Table 2 and then refined the rules using the training set of the gold standard notes.

**Machine Learning-based NLP**

In addition to attempting to automate the annotation of clinical notes with exercise information, several sequence-level binary classification methods were explored to predict whether a specific concept is mentioned in a given span of text at least once according to the gold-standard labels. Here a *sequence* is defined as a string of text within a section that describes an individual exercise. As the therapeutic procedures are documented as numbered lists, it is assumed that each enumerated item that contains text constitutes a single procedure for the purpose of this study. The aim was to extract these procedures from sections and then classify each according to which concepts they mention.

For this task, the sequences provided in the gold standard data were used as raw input, and targets were defined using the labels that were associated with each sequence. These labels consisted of 101 concepts as given by the clinical ontology in Table 2, excluding duration, sets, and reps since these are numeric types unfit for binary classification tasks. Because the post-processed output from MedTagger was formatted in a similar manner to the gold standard data for ease of comparison, a similar method was used to create predictions and directly score MedTagger against the true labels for this task. In this manner we compared our rule-based NLP algorithm against several other methods by redefining the information extraction task as a sequence classification task. The labels of all predicted spans of text were assigned to the section containing it.

Four machine learning models were trained to perform binary classification on sections, including Support Vector Machine (SVM)[14], Logistic Regression (LR)[15], Gradient Boosting[16], and AdaBoost[17]. We built different machine learning models for different physical rehabilitation exercise concept extraction tasks. This resulted in 101 distinct SVM, LR, Gradient Boosting, and AdaBoost models each trained to predict a distinct concept. Each model was created using the scikit-learn[18] library in Python 3. The input for each model was given in a simple uncased bag-of-words vector space fitted to the training set. LR was performed with a learning rate of 1e-4 and balanced class weights. The SVM model used a polynomial kernel with a degree of 2 and also used balanced class weights. AdaBoost and Gradient Boosting were performed with the default parameters provided by scikit-learn, with 100 and 50 estimators, respectively. All unspecified hyperparameters were kept at the default values used by scikit-learn.

**Large Language Model-based NLP**

Recently, Large Language Models (LLMs) have gained much interest due to their promising results across many NLP tasks and straightforward development pipelines. To measure a baseline for the performance of LLMs on this dataset, this study utilized OpenAI's ChatGPT, specifically GPT-3.5-turbo[19], which serves as the NLP backend for the first release of ChatGPT.

This study accessed ChatGPT through the Microsoft Azure secure cloud service with an approved IRB.

For each of the concepts to be identified, a separate prompt was created asking the model to identify whether or not that concept is mentioned in the given sections. We note that the sections used in this exercise were fully de-identified and contained no sensitive information. One of the primary advantages of using LLMs to perform NLP tasks is that they can yield results comparable to other methods with much less time spent on fine-tuning and other manual design tasks. Therefore, in order to evaluate the practicality of using this method in an application with constraints on time or available machine learning expertise, prompts were automatically generated using a template and the name assigned to each label. Two simple prompting techniques are compared: a few-shot prompting technique that included example dialogue for two positive examples and two negative examples[20], and a zero-shot prompt that omitted any examples[21]. Prompts are created by providing a list of dialogue strings that can be attributed to the user providing the prompt, the model giving the response, and a third-party giving instructions to the model, hereafter named System. The following is an example of a few-shot prompt that was automatically generated and sent to ChatGPT:

> System: You are an assistant assigned to determine if a given text segment from a medical record contains mentions of Active Range of Motion. You must answer yes or no.
> User: 7: [PERSONALNAME] maze - AROM LUE - 3 rep
> Model: Yes.
> User: 23: / Supine on mat / dowel rod no wt above head for triceps
> Model: No.
> User: 2: AROM Right wrist ext/flex - 20 x

Model: Yes.

User: 24: SL HS activation- with tapping and suspended to limit friction x10 DEFERRED

Model: No.

User: 14: SLR- 2x10 deferred to HEP

Prompts were made in this manner using four dialogue exchanges, switching out the last line for the section to be classified. Any concepts that were labeled fewer than two times were excluded from the experiment because not enough examples could be generated for the few-shot prompt. Three additional concepts were excluded relating to whether a procedure was performed in-office, at home, or not at all, as well as the three numeric concepts. This left a total of 82 concepts to be extracted.

## Results

Because the clinical ontology covered a broad range of concepts within this domain, many of these concepts were very sparse in our data. More than half of the 101 concepts were present in fewer than ten exercise descriptions in the train or test sets; these concepts have been omitted from the results. Table 4 contains a breakdown of the F1 scores for each machine learning method, as well as the performance of the rule-based NLP algorithm on the NER task, for each of the remaining 40 concepts. See the appendix for the results on all concepts. The best performing machine learning model is shown in bold for each concept.

The rule-based NLP's performance on the sequence classification task was similar to its performance on the NER task. Instances where the former scores higher than the latter can be explained in part by the fact that predicted spans that don't exactly match the corresponding label result in a mismatch for the NER while still resulting in an accurate classification of the section containing the text. The rule-based algorithm tied with or outperformed the other models on half of the concepts in Table 4. Among the machine learning models, Gradient Boosting performed nearly as well, achieving the highest F1 score on 18 concepts.

In addition to these concepts, the rule-based NLP algorithm also predicted the spans of durations, sets, and reps. Since these categories do not have any specific concepts assigned to them, the number present in each span was used instead as a comparison against the true label, converting minutes to seconds where applicable. This resulted in F1 scores of 0.65, 0.58, and 0.88, respectively. It's important to note that we limited the experiments for 'Duration', 'Sets', and 'Reps' exclusively to rule-based algorithms because these categories inherently involve numeric data, which aligns well with the deterministic and pattern-based nature of rule-based approaches.

Gradient Boosting demonstrated the best performance for identifying Range of Motion (ROM) concepts and determining the location of exercise (Performed In-Office, Home Exercise Program, Not Performed) with F1 scores of 0.863 for Active ROM, 0.857 for Active-Assisted ROM, and 0.977, 0.986, 0.950 respectively for the locations. The rule-based Natural Language Processing (RBNLP) algorithm outperformed machine learning models in detecting sides of the body with F1 scores of 0.975 for Right Side and 0.937 for Left Side, and it also performed best on most exercise types, except for Balance/Vestibular and Gait Training concepts, which were classified best by Gradient Boosting with F1 scores of 0.939 and 0.860, respectively. Linear Regression (LR) obtained a strictly higher score than other methods in the weight-bearing exercise concept with an F1 score of 0.876. AdaBoost got a strictly higher score on three concepts, notably on Non-Weight Bearing Positioning with an F1 score of 0.946. The Support Vector Machine (SVM) model did not score higher than other models but had three ties, indicating competitive performance.

These findings indicate that the rule-based approach is particularly effective for certain types of exercises, with superior performance in most categories. However, Gradient Boosting demonstrated strength in more complex categorizations like Balance/Vestibular and Gait Training, where understanding nuanced differences is crucial.

For the LLM-based NLP, results show that both zero-shot prompts and few-shot prompts result in high recall scores that sometimes exceed other methods. However, precision was quite low for most concepts, and F1 scores did not exceed every other method for any concept. However, ChatGPT did occasionally outperform some of the simpler machine learning models, and on one occasion even outperformed the rule-based algorithm (on the backwards plane of motion concept). The average precision over all 82 concepts tested was 0.33 for the zero-shot approach and 0.27 for the few-shot approach. The average recall was 0.8 for zero-shot and 0.82 for few-shot. This resulted in average F1 scores of 0.37 and 0.35, respectively, indicating that the zero-shot approach was slightly better on average than the few-shot approach. However, the few-shot approach performed the best for all but ten concepts. The reason the zero-shot method performed better on average is thus due to the fact that it shows significant improvement on a few specific concepts, such as hip, scapula, hand, abduction and extension.

Table 4: Binary F1 scores of each algorithm on the test set (50 documents total). (RBNLP: Rule-based NLP, LR: Linear Regression, SVM: Support Vector Machine, NER: Named Entity Recognition)

| Category | Concept | RBNLP NER | RBNLP Sequence | LR | SVM | AdaBoost | Gradient Boosting | ChatGPT (few-shot) | ChatGPT (zero-shot) | Training Set Size | Test Set Size |
|---|---|---|---|---|---|---|---|---|---|---|---|
| Description | Performed In-Office | 0.957 | 0.976 | 0.970 | 0.960 | 0.977 | **0.983** | N/A | N/A | 2464 | 497 |
| | Home Exercise Program | **0.986** | **0.986** | **0.986** | 0.938 | **0.986** | **0.986** | N/A | N/A | 93 | 34 |
| | Not Performed | 0.949 | 0.949 | 0.923 | 0.909 | 0.936 | **0.950** | N/A | N/A | 1295 | 206 |
| ROM | Active | 0.839 | 0.830 | 0.824 | 0.840 | **0.863** | **0.863** | 0.321 | 0.109 | 103 | 22 |
| | Active-Assisted | 0.769 | 0.769 | 0.800 | 0.791 | 0.837 | **0.857** | 0.543 | 0.210 | 160 | 24 |
| | Passive | 0.952 | 0.938 | **0.970** | 0.903 | 0.938 | **0.970** | 0.552 | 0.198 | 121 | 16 |
| Side | Right Side | 0.912 | **0.975** | 0.674 | 0.851 | 0.628 | 0.680 | 0.912 | 0.878 | 548 | 97 |
| | Left Side | 0.912 | **0.937** | 0.763 | 0.823 | 0.721 | 0.752 | 0.823 | 0.832 | 462 | 134 |
| | Bilateral | 0.772 | **0.907** | 0.559 | 0.474 | 0.667 | 0.659 | 0.706 | 0.723 | 260 | 51 |
| Location | Upper Extremity | 0.847 | **0.939** | 0.879 | 0.847 | 0.901 | 0.876 | 0.291 | 0.241 | 285 | 47 |
| | Lower Extremity | 0.955 | 0.936 | 0.936 | 0.930 | 0.966 | **0.978** | 0.378 | 0.339 | 223 | 44 |
| | Hip | 0.949 | 0.947 | **0.973** | **0.973** | 0.943 | 0.972 | 0.403 | 0.806 | 168 | 36 |
| | Knee | 0.950 | 0.950 | 0.919 | 0.882 | **0.974** | **0.974** | 0.469 | 0.434 | 108 | 19 |
| | Ankle | **1.000** | **1.000** | 0.923 | 0.600 | **1.000** | **1.000** | 0.607 | 0.262 | 55 | 14 |
| | Shoulder | 0.936 | **0.977** | 0.952 | 0.952 | 0.953 | 0.953 | 0.744 | 0.548 | 224 | 44 |
| | Scapula | **0.833** | **0.833** | 0.783 | 0.700 | **0.833** | **0.833** | 0.525 | 0.607 | 72 | 10 |
|  | Elbow | **0.967** | 0.963 | 0.963 | 0.943 | 0.923 | 0.923 | 0.848 | 0.447 | 147 | 26 |
|  | Forearm | 0.815 | 0.833 | 0.870 | **0.952** | 0.870 | **0.952** | 0.151 | 0.204 | 86 | 10 |
|  | Wrist | **0.902** | 0.898 | 0.826 | 0.773 | 0.875 | 0.875 | 0.600 | 0.314 | 129 | 23 |
|  | Hand | **0.951** | 0.944 | 0.926 | 0.848 | 0.925 | 0.949 | 0.438 | 0.574 | 243 | 68 |
| Plane | Abduction | 0.976 | **0.985** | 0.971 | 0.937 | 0.971 | 0.971 | 0.576 | 0.839 | 170 | 33 |
|  | Anterior | 0.545 | 0.545 | **0.750** | 0.667 | **0.750** | 0.667 | 0.221 | 0.195 | 22 | 10 |
|  | Backward | 0.727 | 0.720 | 0.688 | 0.800 | **0.952** | 0.846 | 0.720 | 0.790 | 92 | 11 |
|  | Extension | 0.980 | 0.980 | 0.979 | 0.933 | **0.989** | **0.989** | 0.556 | 0.684 | 266 | 48 |
|  | External Rotation | 0.897 | **0.917** | 0.870 | 0.818 | 0.870 | 0.870 | 0.655 | 0.543 | 74 | 11 |
|  | Flexion | 0.956 | 0.947 | **0.964** | 0.955 | **0.964** | **0.964** | 0.757 | 0.615 | 327 | 55 |
|  | Forward | **0.977** | 0.974 | 0.857 | 0.865 | 0.950 | 0.900 | 0.667 | 0.729 | 148 | 19 |
|  | Lateral | 0.577 | 0.588 | 0.786 | 0.837 | **0.870** | 0.851 | 0.546 | 0.373 | 132 | 23 |
|  | Supination | **0.923** | 0.917 | 0.880 | 0.917 | 0.917 | 0.917 | 0.550 | 0.480 | 82 | 11 |
| Exercise Type | Upper Extremity Strength | **0.913** | **0.913** | 0.840 | 0.791 | **0.913** | 0.894 | 0.272 | 0.166 | 138 | 21 |
|  | Lower Extremity Strength | 0.926 | **0.969** | 0.913 | 0.894 | 0.924 | 0.894 | 0.449 | 0.332 | 447 | 97 |
|  | Trunk/Core Strength | 0.897 | **0.889** | 0.692 | 0.471 | 0.471 | 0.700 | 0.104 | 0.090 | 35 | 12 |
|  | Range of Motion | 0.853 | **0.876** | 0.842 | 0.843 | 0.725 | 0.674 | 0.301 | 0.153 | 257 | 53 |
|  | Flexibility/Mobility | 0.962 | **0.974** | 0.909 | 0.857 | 0.947 | 0.949 | 0.279 | 0.147 | 178 | 38 |
|  | Balance/Vestibular | 0.787 | 0.752 | 0.852 | 0.809 | 0.882 | **0.939** | 0.597 | 0.470 | 351 | 47 |
| | Gait Training | 0.808 | 0.837 | 0.837 | 0.814 | 0.851 | **0.860** | 0.626 | 0.529 | 310 | 47 |
| | Functional Mobility | 0.775 | **0.831** | 0.727 | 0.750 | 0.691 | 0.780 | 0.220 | 0.182 | 204 | 33 |
| Purpose | Simulated | 0.769 | 0.769 | **0.870** | 0.762 | 0.857 | **0.870** | 0.688 | 0.667 | 48 | 10 |
| | Weight Bearing | 0.788 | 0.833 | **0.876** | 0.867 | 0.857 | 0.871 | 0.197 | 0.282 | 255 | 43 |
| Positioning | Non-Weight Bearing | 0.931 | 0.932 | 0.916 | 0.918 | **0.946** | 0.923 | 0.283 | 0.038 | 539 | 91 |
| Average | | 0.878 | **0.891** | 0.861 | 0.835 | 0.875 | 0.883 | 0.502 | 0.433 | 283 | 53 |

# Discussion

As indicated by the high performance of the machine learning models on many of the concepts, the task of extracting information from exercise descriptions was not complex. Although some of these concepts could be extracted effectively using straightforward rules or a small machine learning model, there were also many cases where clinical notes appear ambiguous without context. For instance, the abbreviation "SL" could be interpreted as "single leg" or "side-lying" depending on the exercise being described. In addition, "L" Could mean "left" or "lateral", which explains why the rule-based NLP algorithm performed slightly worse when classifying left vs right. Single letters being used as abbreviations, especially "A" as a shorthand for "anterior", could cause issues in machine learning algorithms without careful consideration. It would be possible to increase the performance of the rule-based algorithm by further tuning the rules to search for context clues at other points in the document, but this could potentially cause the rules to overfit the training set. Of particular interest are the numeric data present in duration, sets, and reps. These are particularly tricky to extract since they are expressed in a wide variety of ways by different physicians. It can be difficult to define what sets and reps are depending on the exercise, and sometimes one or both are not well-defined at all. Additionally, the use of apostrophes and quotes can either indicate measurements of time or distance, once again requiring context to disambiguate. Mentions of distance were not annotated in the gold-standard labels, but it is important in measuring the intensity of some exercises, so we plan to include it in the future.

Some of the misclassifications of the rule-based algorithm are due to inaccuracies in the gold-standard dataset. For instance, many false positives produced by the rule-based algorithm appeared to be concepts that were missed by the annotators. There were also a few minor errors that could be explained by a mouse slip, including a span of text being assigned the wrong concept or a span excluding the last letter in a word. There were also some spelling mistakes in the notes themselves; common instances were explicitly mentioned in the rules to increase precision. Preprocessing clinical notes to correct spelling mistakes might be useful to improve results, though this creates a risk of incorrect changes being made to uncommon words. All of these errors were not particularly common throughout the labels, but they could have a significant effect on concepts that are already uncommon in the data.

Another obstacle that obscured some of the signal in the data came from the de-identification process. In addition to removing names, addresses, and other protected information from these documents, many other tokens and phrases were mistakenly removed, including equipment names and numbers denoting indices in a list. These were replaced with placeholder tokens such as "[ADDRESS]" or "[PERSONALNAME]". The low precision of the de-identification process caused some relevant information to be obfuscated or entirely erased from notes.

During the data annotation, we found that many of the concepts identified as relevant in this domain were not well documented in the data we extracted for annotation. This could be due in part to the fact that the data was only collected from stroke patients, but this is not expected to be the main reason because stroke patients can have a wide variety of musculoskeletal problems, resulting in a correspondingly wide variety of treatments being mentioned in clinical notes[22]. The other reason the dataset lacks many of these concepts could be that they are rarely mentioned in these particular sections of clinical notes, either because they are not common enough to appear in many records at all or because they are mentioned more often in other sections. Thus, future research could focus on improving extraction methods to focus more on these uncommon concepts or include information from outside of the exercise descriptions.

In addition to ChatGPT for the LLM-based NLP approach, we also fine-tuned a BERT model with the task of categorizing physical rehabilitation exercise concept. The BioClinical BERT model[23] was used, which was pretrained on MIMIC-III[24]. However, the amount of data collected seemed insufficient to make the model perform comparably to simpler methods. The model with the highest F1 score on the validation set had an average F1 score of 0.05 across all concepts on the test set. It was able to predict whether an exercise was performed in-office with an F1 score of 0.72, but all other 100 other concepts scored between 0 and 0.35. Therefore, we didn't include this approach in the experimental comparison.

**Limitations and Future Work:**

One limitation in our current research was the necessary exclusion of 'Duration', 'Number of Sets', and 'Number of Reps' from our machine learning-based NLP analysis due to their numeric nature, rendering them unsuitable for binary classification tasks. In the future work, we plan to incorporate regression models or specialized classification techniques capable of handling numeric data. We also plan to expand our research to include additional variables such as stroke duration and severity, recognizing their potential to significantly enhance the prediction accuracy and effectiveness of rehabilitation strategies.

Additionally, we acknowledge the discrepancy between the use of technique names and specific motion types in rehabilitation exercises. To address this, we intend to develop a supplementary module for our algorithm that effectively maps popular technique names to their corresponding motion types and categories, enhancing the algorithm's comprehensiveness and applicability.

Moreover, we plan to implement a robust standardized extraction protocol in the next version of our algorithm to mitigate the omission of therapeutic procedure sections due to copy-and-paste errors. This protocol will include multiple checks for consistency and completeness and will be assessed through a pilot study to ensure its reliability and accuracy. To enhance our model's generalizability amidst varied note-writing practices across rehabilitation facilities, future research will also focus on diversifying data sources, refining adaptability to diverse writing styles and terminologies, and conducting extensive validation studies in a range of settings to improve performance. Through continuous monitoring and refinement of our extraction process, we are committed to enhancing the reliability and validity of our data, thereby strengthening the overall quality and impact of our research.

## Conclusion

In this study, we developed and evaluated several NLP algorithms to extract physical rehabilitation exercise information from clinical notes of post-stroke patients. We first created a novel and comprehensive clinical ontology to represent physical rehabilitation exercise in clinical notes, and then developed a variety of NLP algorithms leveraging the state-of-the-art techniques, including rule-based NLP algorithms, machine learning-based NLP algorithms, and LLM-based NLP algorithms. The experiments on the clinical notes of a cohort of post-stroke patients showed that the rule-based NLP algorithm had the best performance for most of the physical rehabilitation exercise concepts. Among all machine learning models, Gradient Boosting achieved the best performance on a majority of concepts. On the other hand, the rule-based NLP performed well for extracting handled durations, sets, and reps while Gradient Boosting excelled in ROM, location detection. LLM-based NLP achieved high recall with zero-shot and few-shot prompts but low precision and F1 scores. It occasionally outperformed simpler models and once beat the rule-based algorithm.

## Acknowledgement

This work was supported by the School of Health and Rehabilitation Sciences Dean's Research and Development Award.

## Conflicts of Interest

None declared.

## Abbreviations

NLP: natural language processing

LLM: large language model

EHR: electronic health record

IRB: Institutional Review Board

NER: named entity recognition

SVM: Support Vector Machine

LR: Logistic Regression